\documentclass[11pt,a4paper]{amsart}

\usepackage[utf8]{inputenc}
\usepackage[T1]{fontenc}
\usepackage[english]{babel}
\usepackage{amsmath, amssymb, amsthm}
\usepackage{graphicx}
\usepackage{hyperref}
\usepackage{geometry}
\usepackage{enumitem}
\usepackage{color}
\usepackage{float}
\usepackage{booktabs}
\usepackage{multirow}
\usepackage{caption}
\usepackage{subcaption}

\title{Graph Neural Networks in Wind Power Forecasting}
\author{Javier Castellano}

\address{Ravenwits}

\email{javier.castellano@ravenwits.com}

\author{Ignacio Villanueva}

\address{Ravenwits \& Departamento de An\'alisis Matem\'atico \\
Facultad de Matem\'aticas \\ Universidad Complutense de Madrid \\
Madrid}

\email{ignaciov@mat.ucm.es}

\thanks{Partially funded by {\it Subvenciones dirigidas al desarrollo de casos de uso de inteligencia artificial aplicada a la industria de la Comunidad Autónoma de Madrid} 
03-ACU1-00013.1/2024. Financed by NextGenerationEU}

\begin{document}

\begin{abstract}
We study the applicability of GNNs to the problem of wind energy forecasting. We find that certain architectures achieve performance comparable to our best CNN-based benchmark.
The study is conducted on three wind power facilities using five years of historical data. Numerical Weather Prediction (NWP) variables were used as predictors, and models were evaluated on a 24–36 hour ahead test horizon.
\end{abstract}



\maketitle


\section{Introduction}\label{S:introduction}

Accurate forecasting of renewable energy production, particularly wind power, is a key enabler for the reliable and efficient operation of modern power systems. As the share of renewable energy sources increases, so does the need for short- to medium-term accurate forecasts. 

Traditionally, classical statistical methods and physical models have dominated this field, but in recent years, machine learning approaches—especially those based on deep learning—have shown promising performance improvements.

Convolutional Neural Networks (CNNs) have been particularly effective in this task \cite{Wa, Li}. When applied to Numerical Weather Prediction (NWP) they seem to be able to capture mesoscale weather patterns. However, CNNs inherently have translational invariance properties, which are not intrinsic to the problem. This fact motivates the exploration of alternative models. 

Graph Neural Networks (GNNs) provide a powerful framework for learning from graph-structured data and have recently gained attention in the meteorological community. See \cite{GC} and the references therein. Notably, the GraphCast project by Deep Mind \cite{GC} has demonstrated state-of-the-art performance in global weather forecasting using GNNs. The heuristics behind this success are that GNN structure should be useful when learning complex time and space dependencies. In particular, their structure should be well suited to problems ruled by an underlying differential equation \cite{Ba}.

Up to now, the application of GNNs in wind energy prediction has primarily focused on modeling wind farms as graphs, where each node corresponds to a turbine. This representation allows the network to capture spatial relationships and interactions among turbines, leveraging the graph structure \cite{Da}, outperforming other approaches such as LightGBM \cite{lgbm}. We hypothesize that GNNs should also be effective in capturing complex relationships between meteorological variables, given that GNNs can be viewed as a generalization of CNNs \cite{He}. This, combined with the strong baseline performance typically observed with CNNs \cite{Wa, Li}, further motivates their use in this context.

In this work, we investigate the applicability of GNNs to forecasting wind energy production. We experiment with various GNN architectures and input features, including NWP data, topographical variables, and historical production. Our results show that GNN-based models can match the performance of state-of-the-art CNNs on the same prediction tasks, highlighting their potential as a flexible and powerful alternative for forecasting in energy systems.

The structure of this note is as follows: 
In Section \ref{S:results} we present the results of our experiments, comparing the best GNN architectures we have obtained with our CNN benchmarks for the same plants and training and test periods. 
In Section \ref{S:data} we describe the data, both NWP and production, which we have used for our experiments. 
In Section \ref{S:GNN} we describe the GNN architectures we have tested in our work. 
Finally, in Section \ref{S:conclusions} we present the conclusions of our results and the lines of work along these ideas that we want to study in the near future.

\section{Results}\label{S:results}
We present an evaluation of our proposed GNN model compared to the best CNN baseline for wind energy forecasting that we use in Ravenwits. The evaluation was carried out on several wind farms located in Romania, as shown in Table \ref{tab:comparation}.

The evaluation strategy consisted of training the models on data from four consecutive years and testing on data from a separate fifth year. By employing two different train-test partitions, we obtain a more robust estimate of model performance, due to the variability in error magnitude across different test years. Moreover, the errors correspond to using the average of two identical runs of the same model as the prediction, in order to obtain performance metrics with lower variance.

\begin{table}[ht]
    \centering
    \begin{tabular}{@{} l r r r r r@{}} 
        \toprule
        Test Year& 2022 &  2021 & 2022 & 2021 & 2023\\
        Wind farm & A & A & B & B & C\\
        \midrule
        CNN (MAE)       & 8.16 & \textbf{\small8.15} & 11.55 & 12.14 & 10.90\\
        GNN (MAE)      & \textbf{\small8.10} & 8.23 & \textbf{\small11.49} & \textbf{\small12.12} & \textbf{\small10.83}\\
        \bottomrule
    \end{tabular}
    \caption{Mean Absolute Error (MAE) results for different wind farms and test years with a forecasting horizon ranging from 24 to 36 hours. The predicted and target values range between 0 and 100, representing energy production as a percentage of the installed capacity. This normalization allows the magnitude of the errors to be comparable across wind farms. Wind farms are anonymized using letters instead of their real names.}
    \label{tab:comparation}
\end{table}

The architecture and hyperparameters of the CNN model used in this study were pre-established based on previous research, ensuring a strong and well-optimized baseline. In contrast, the hyperparameters for our GNN model were meticulously selected using a separate wind farm for validation. More detailed information on the hyperparameter optimization process can be found in Section \ref{S:GNN} of this article.

The errors presented in Table \ref{tab:comparation} indicate that the two proposed models show comparable accuracy. Minor variations across different scenarios, where one model slightly outperforms the other and vice versa, support the conclusion that both models perform quite similar.

\section{Data}\label{S:data}
\subsection{Labels}
The objective is to predict the average aggregated power output of a group of wind turbines in a wind farm over a given time interval, typically an hourly one. Accordingly, for each wind farm, historical production data is available, where for each hour $h$, a scalar value represents the average aggregated power generated between hour $h-1$ and $h$.

This production metric aims to reflect a measure that is independent of external factors such as turbine unavailability or TSO limitations. So, it is a quantity that is physically related to meteorological conditions. In other words, the goal is to predict the total power obtainable: the expected power output of the wind farm assuming it is unaffected by external factors unrelated to the physical and meteorological dynamics of the problem.

\subsection{Numerical weather prediction (NWP)}

The explanatory variables consist of Numerical Weather Predictions (NWPs). They are the output of the solution to the models run by the meteorological agencies. These models simulate atmospheric evolution by solving differential equations based on the laws of physics. These forecasts include the horizontal $u$ and $v$ components of wind at several altitudes, radiation, cloud cover, temperature, and other relevant meteorological variables. In this case, we focus on the components of wind at 100 meters above the surface which is the typical height of wind turbines.

The model output is provided on a latitude-longitude grid, with temporal resolution typically at one-hour intervals. Spatial resolution varies by provider. In this study we employ the European Centre for Medium-Range Weather Forecasts’ (ECMWF) High-Resolution (HRES) model, which offers a grid spacing of $0.1^\circ$ (approximately 11 km at the equator). Therefore, for each latitude and longitude, the forecast for a given hour consists of the components of the wind at 100 meters at that location.

For the CNN-based models, a $2^\circ \times 2^\circ$ ($40 \times 40$) map centered on the wind farm is used, whereas for the GNN-based models, a smaller $1^\circ \times 1^\circ$ ($20 \times 20$) map is employed. The choice of a smaller map for the GNN architecture is primarily motivated by computational cost, as it reduces the number of nodes to be processed in the GNN layers by a factor of four.

\section{Graph Neural Networks}\label{S:GNN}

Graph Neural Networks (GNNs) have emerged as a powerful class of deep learning models for structured data, particularly suited for representing complex physical systems. Notably, models such as \textit{GraphCast} \cite{GC} have demonstrated the effectiveness of GNNs in high-impact applications like weather forecasting, where they model the spatiotemporal dynamics of atmospheric variables. In general, GNNs have also shown promise in solving partial differential equations and other physics-informed learning tasks, motivating their use for wind power forecast, which involves underlying physical processes. See \cite{Ba} and the references therein. 

A GNN operates on a graph input, denoted by $ G = (V, E) $, where \( V \) is the set of nodes and $ E \subseteq V \times V $ the set of edges, which may be directed. Each node $ v_i \in V $ is associated with a feature vector so we can see it as $v_i\in\mathbb{R}^{d_{node}}$; in our case, $d_{node}=2$ containing the horizontal wind components $ (u, v) $ at 100 meters above ground level. Similarly, each edge or arc $ e_{ij} \in E $, representing a connection from node $ v_i $ to node $ v_j $, may also carry features, such as distance or directionality, depending on the nature of the graph.

GNNs are fundamentally characterized by their ability to aggregate and transform information from a node's neighbors through a process known as \textit{message passing}. As information is passed through the layers, each node updates its representation by combining its own features with those of its neighbors, allowing the model to learn local and global patterns on the graph. The highly customizable way of performing \textit{message passing} makes them very versatile, allowing to capture detailed biases of the problem in the architecture itself. There are various types of relevant GNN layers that implement different forms of message propagation, such as Graph Convolutional Networks \cite{Ki} or Graph Attention Networks \cite{Br, Ve}.

\subsection{Architecture}

In our work, we adopt a message passing scheme inspired by the \textit{processor} block in \cite{GC}. A single message passing layer consists of the following steps:

\begin{enumerate}[label=\arabic*.]
    \item For each arc\footnote{Although \( e_{ij} \in E \Leftrightarrow e_{ji} \in E \) in our problem, the graph is considered directed because the corresponding arc features differ, as explained in Section~\ref{S:GNN-input}.} $e_{ij}$ from node $v_i$ to node $v_j$, we apply a differentiable parametric function
    \begin{equation}\label{eq:1}
        f: \mathbb{R}^d \times \mathbb{R}^d \times \mathbb{R}^e \rightarrow \mathbb{R}^{\hat{e}},
    \end{equation}
    where $d$ and $e$ denote the dimensionality of the node and arc features, respectively. The function takes as input the source node features $v_i$, the target node features $v_j$, and the edge features \( e_{ij} \), and produces a \textit{message} vector. Besides, these new vectors will be the new arc features for the consecutive layers.
    
    \item For each node $v_i$, incoming messages are aggregated using a \textit{permutation-invariant} operator, typically a summation:
    \begin{equation}\label{eq:2}
        m_i = \sum_{j : (j,i) \in E} f(v_j, v_i, e_{ji}).
    \end{equation}
    
    \item The node features are then updated via another differentiable function
    \begin{equation}\label{eq:3}
        g: \mathbb{R}^d \times \mathbb{R}^{\hat{e}} \rightarrow \mathbb{R}^{\hat{d}}
    \end{equation}
    which outputs the new node representation:
    \[
    v_i' = g(v_i, m_i).
    \]
\end{enumerate}

This modular design allows for expressive modeling of spatial dependencies and physical interactions between locations.

At the first layer, we use an input dimensionality of $d=2$ for the nodes (wind components) and $e=3$ for the edge features, as defined in \ref{S:GNN-input}. After this layer, both the node and edge dimensions ($\hat{d}$ and $\hat{e}$) for the middle layers become hyperparameters of the model.

To predict the scalar value of wind power generation, we set $\hat{d}=1$ in the final graph layer, effectively compressing each node representation to a single feature. These node outputs are then aggregated using a final dense layer with a single unit acting as a linear regressor. Thus, the GNN layers can be interpreted as a hierarchical feature extraction mechanism, where each node encodes its contribution to the overall power output.

\subsection{Input graph}\label{S:GNN-input}
The input to the model consists of the weather map at the corresponding time, represented as a tensor of shape $(20, 20, 2)$. The two channels vectors, at each node, correspond to the components of the wind.

The adjacency matrix can be constructed in several ways; in this work, we adopted a simple approach: nodes are considered adjacent if they are neighbors in the same row or in the same column. In addition, edges are added to connect nodes with others located at a fixed distance $x$ along the same row or column. This design allows for interaction between variables at a given geographical point and more distant ones, without the need for multiple layers of message passing.

\begin{figure}[htbp]
  \centering
  \includegraphics[width=0.7\textwidth]{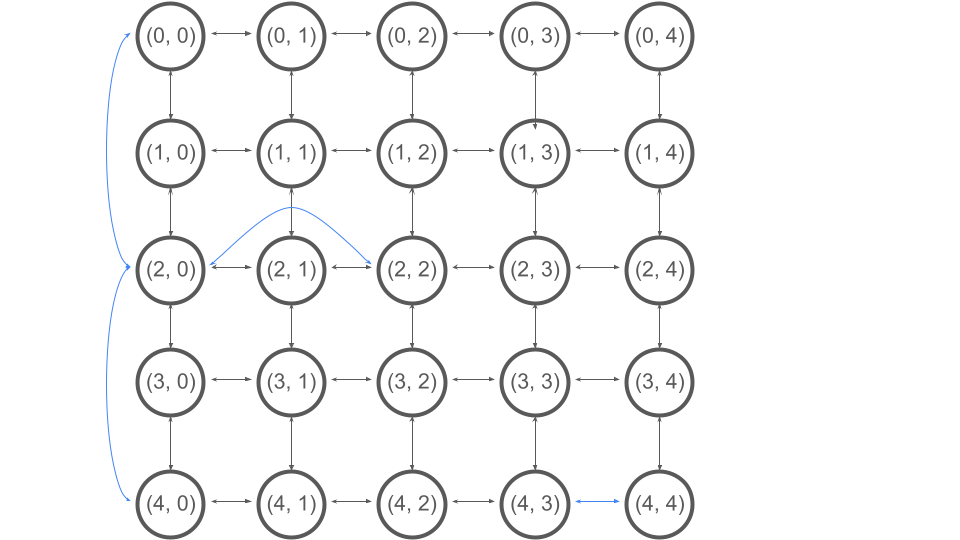}
  \caption{Example of an input graph for the GNN, where each node represents the wind at a specific geographic location defined by its latitude and longitude.
Nodes are connected to their neighboring locations (black arrows).
For clarity, only the distant connections from node $(2, 0)$ are shown in blue, to maintain the readability of the graph.}
  \label{F:1}
\end{figure}

A graphical example using a $5\times 5$ map with extra edges that skip 2 nodes is shown in Figure \ref{F:1}.

Finally, we define the edge features. For this, if node $v_i$ is located at coordinates $(a, b)$, $a, b \in\{0, 1, \dots, 19\}$, and node $v_j$ at $(a', b')$, then the arc feature vector $e_{ij}$ is defined as:

\begin{equation}\label{eq:4}
    e_{ij} = \left( a'-a, b'-b, \sqrt{(a'-a)^2 + (b'-b)^2} \right) \in \mathbb{R}^3
\end{equation}

This formulation enables the graph neural network to handle messages in a more specialized  way by providing information about both the direction and distance of each connection. It captures the idea that messages from distant nodes may not need to be treated in the same way as those from nearby ones. Also, this approach allows nodes to send information other than that received from another node. It is important to note that if $e_{ij} = e_{ji}$, the model would treat messages sent in both directions identically (see \ref{eq:2}), which could limit the network’s expressive capacity.

\subsection{Details}\label{S:GNN-details}
Following the same approach as in \cite{GC}, the functions used in \ref{eq:1} and \ref{eq:3} consist of a two-layer MLP with Sigmoid Linear Unit (SiLU) activation applied at each layer. The use of two layers for processing nodes and edges plays a fundamental role compared to other approaches such as \cite{Ki}, as it allows for non-linearity in the interactions between nodes and edges.

The number of layers in the GNN can depend on the amount of available data, with a reasonable range being between 2 and 10 layers. The optimizer used was Adam, combined with a learning rate schedule employing exponential decay with a factor of 0.98 after each epoch. A validation set was used for early stopping to avoid overfitting. The batch size was set to $64$.

L1 regularization was applied to the final dense layer responsible for predicting the production, encouraging sparsity by driving to zero the weights associated with non-informative nodes, thereby reducing potential overfitting.

Regarding implementation, the model was developed using the PyTorch library along with the \textit{torch-scatter} package, which enables efficient parallelization of our custom message-passing layers. Training was conducted on an NVIDIA A100 GPU with 80 GB of VRAM in the LOVELACE cluster (CSIC-ICMAT). This setup allowed for reduced data transfer times between RAM and VRAM. Each training epoch took approximately 45 seconds, with convergence typically reached in around 50 epochs.

\subsection{Explored variants and hyperparameters}\label{S:arquitectures}
Hyperparameter selection was performed using a validation set based on a wind farm different from those presented in Table \ref{tab:comparation}. The explored hyperparameters included: the number of layers, the dimensionality of the latent space for nodes and edges, L1 regularization and dropout rates, the inclusion of an additional dense layer before the final prediction layer, and the learning rate schedule.

\section{Conclusions and future research}\label{S:conclusions}

We have shown that GNNs can be as accurate as CNNs in short- and medium-term wind energy production forecasting. Since GNNs have only recently started to be used for this task, it is likely that their use can still be refined and that they may soon outperform CNNs.

We believe it is worth continuing to pursue this direction, exploring new architectures capable of learning meteorological behavior even more effectively.

In particular, we believe that the exploration of different message passing mechanisms should help in this direction.

\end{document}